\documentclass{article}

\pdfoutput=1

\usepackage[utf8]{inputenc} 
\usepackage[T1]{fontenc}    
\usepackage{hyperref}       
\usepackage{url}            
\usepackage{booktabs}       
\usepackage{amsfonts}       
\usepackage{nicefrac}       
\usepackage{microtype}      
\usepackage{xcolor}         

\usepackage{algpseudocode}
\usepackage{microtype}
\usepackage{graphicx}
\usepackage{subfigure}
\usepackage{multirow}
\usepackage{array}
\usepackage{colortbl}
\usepackage{tabularx}         
\usepackage{adjustbox}        
\usepackage{arydshln}
\usepackage{xspace}
\usepackage{algorithm}
\usepackage{enumitem}
\usepackage[numbers, sort&compress]{natbib}

\usepackage{arxiv}

\usepackage{amsmath}
\usepackage{amssymb}
\usepackage{mathtools}
\usepackage{amsthm}

\theoremstyle{plain}
\newtheorem{theorem}{Theorem}[section]

\newtheorem{lemma}[theorem]{Lemma}
\newtheorem{corollary}[theorem]{Corollary}
\theoremstyle{definition}

\newtheorem{assumption}[theorem]{Assumption}
\theoremstyle{remark}

\definecolor{skyblue}{RGB}{229,242,247}
\definecolor{blue}{RGB}{66, 109, 181}

\newcommand{\ie}{\emph{i.e.}\xspace}
\newcommand{\eg}{\emph{e.g.}\xspace}
\newcommand{\celora}{{CE-LoRA}\xspace}
\newcommand{\textub}[1]{\underline{\textbf{#1}}}

\newcommand{\blfootnote}[1]{%
  \begingroup
  \renewcommand{\thefootnote}{}
  \footnotetext{#1}
  \addtocounter{footnote}{-1}%
  \endgroup
}

\usepackage{abstract}

\title{\celora: Computation-Efficient LoRA \\ Fine-Tuning for Language Models}

\author{%
  Guanduo~Chen$^{* \dag}$\\
  Fudan University\\
  \texttt{gdchen22@m.fudan.edu.cn} \\
  \And
  Yutong~He$^{\dag}$\\
  Peking University\\
  \texttt{yutonghe@pku.edu.cn} \\
  \And
  Yipeng~Hu\\
  Peking University\\
  \texttt{yipenghu@pku.edu.cn} \\
  \And
  Kun~Yuan$^\ddagger$\\
  Peking University\\
  \texttt{kunyuan@pku.edu.cn} \\
  \And
  Binhang Yuan$^\ddagger$\\
  HKUST \\
  \texttt{biyuan@ust.hk}\\
}

\begin{document}
\maketitle

\blfootnote{$^*$ Work done when the author was working as a research assistant under the supervision of Binhang Yuan.}
\blfootnote{$^\dag$ Both authors contributed equally to this research.}
\blfootnote{ $^\dagger$ Coressponding author.}

\begin{abstract}
Large Language Models (LLMs) demonstrate exceptional performance across various tasks but demand substantial computational resources even for fine-tuning computation. Although Low-Rank Adaptation (LoRA) significantly alleviates memory consumption during fine-tuning, its impact on computational cost reduction is limited. This paper identifies the computation of activation gradients as the primary bottleneck in LoRA's backward propagation and introduces the \underline{\textbf{C}}omputation-\underline{\textbf{E}}fficient \underline{\textbf{LoRA}} (\textbf{CE-LoRA}) algorithm, which enhances computational efficiency while preserving memory efficiency. \celora leverages two key techniques: Approximated Matrix Multiplication, which replaces dense multiplications of large and complete matrices with sparse multiplications involving only critical rows and columns, and the Double-LoRA technique, which  reduces error propagation in activation gradients. Theoretically, \celora converges at the same rate as LoRA, \( \mathcal{O}(1/\sqrt{T}) \), where $T$ is the number of iterations. Empirical evaluations confirm that \celora significantly reduces computational costs compared to LoRA without notable performance degradation.
\end{abstract}

\vspace{-1em}
\section{Introduction}

Large Language Models (LLMs) have garnered significant attention in recent years for their exceptional performance across a wide range of practical tasks, including machine translation, commonsense reasoning, and planning, among others~\cite{bommasani2021opportunities}.
The versatility of these models has also driven a growing demand for fine-tuning them on specific tasks or domains to unlock their full potential~\cite{zhang2023instruction,han2024parameter}.
However, fine-tuning these models remains a highly resource-intensive process, demanding substantial computational power and large amounts of GPU memory. As model parameters and training tokens scale up, the increasing training costs have made it difficult for most organizations to keep pace with advancements in LLM research due to resource constraints.

To tackle these challenges, recent advancements such as Low-Rank Adaptation (LoRA) \citep{hu2021lora} have demonstrated promising results in reducing memory consumption during LLM fine-tuning, enabling the fine-tuning of models with more parameters or larger batch sizes within constrained resources. While LoRA significantly alleviates memory requirements, its reduction of computational costs remains limited --- although the low-rank adapters can save part of the computation costs by reducing the matrix sizes, the size of original weight matrices used to calculate \textit{activation gradients} remains unchanged, which contributes to half of the total computation cost in the backpropagation of the original model. The formulation of this computation is illustrated in Section~\ref{sec:preliminaries}. Limited by this computational bottleneck, LoRA can reduce computation by at most half during the backpropagation process. This limitation raises the following open question:
\begin{quote}
    \textit{Compared with vanilla LoRA, can we develop a more computation-efficient fine-tune algorithm by the same memory budget without sacrificing the statistical efficiency (i.e., convergence)?}
\end{quote}

To answer this question, we first conduct a computational analysis of LoRA's backward propagation procedure and identify the primary computational bottleneck as the \textit{calculation of the activation gradients}. This step accounts for the majority of the backward computation load, especially when LoRA employs a relatively small rank $r$.





\begin{figure*}[!t]
  \centering
  \begin{tabular}{@{}c@{\hspace{0.05\textwidth}}c@{}}  
    \includegraphics[width=0.425\textwidth]{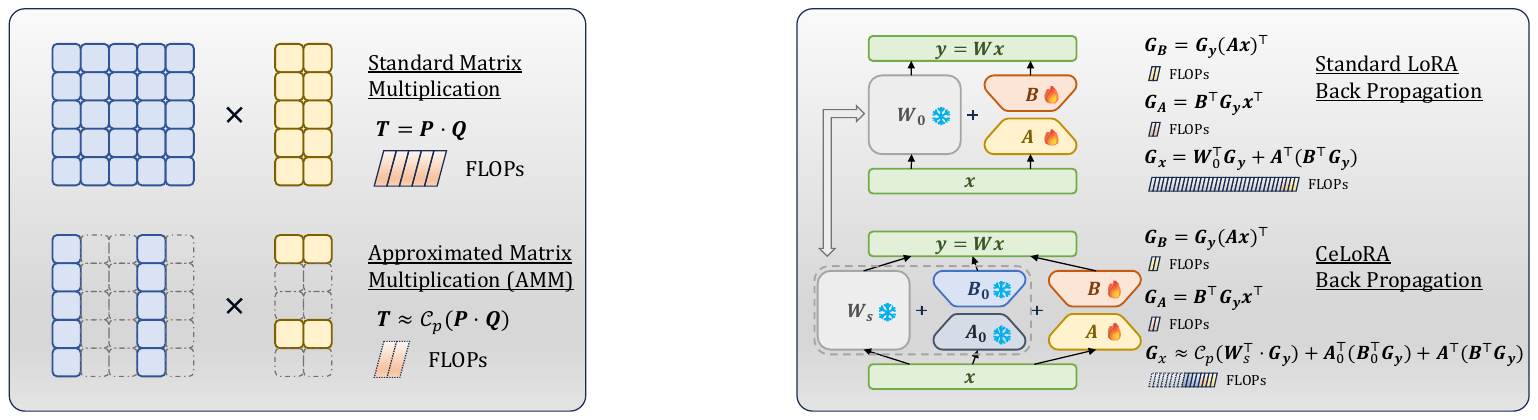} &
    \includegraphics[width=0.540\textwidth]{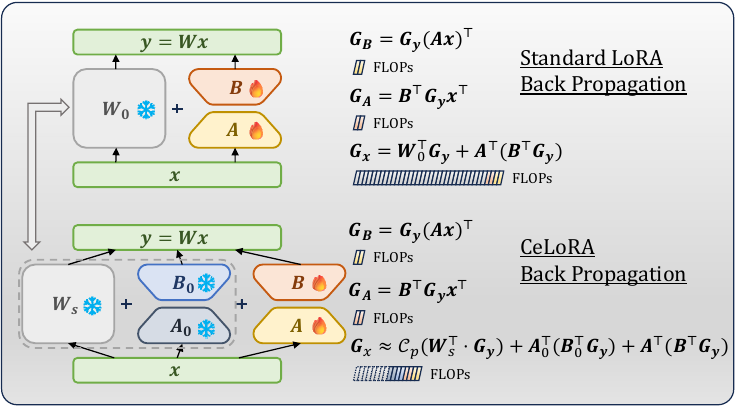} \\
  \end{tabular}
  \caption{\small An illustration of the Approximated Matrix Multiplication (AMM) technique (left) and the \celora framework (right).}
  \label{fig:intro-draft}
  \vspace{-1em}
\end{figure*}

Based on this analysis, we propose the Approximated Matrix Multiplication (AMM) technique to reduce the computation of activation gradients. In order to reduce the computation of a dense matrix multiplication $\mathbf{T}=\mathbf{P}\mathbf{Q}$, where $\mathbf{P}\in\mathbb{R}^{m\times n}$ and $\mathbf{Q}\in\mathbb{R}^{n\times k}$, AMM directly reduces the size of the matrices by discarding unimportant rows or columns, obtaining $\mathbf{P}[:,\mathcal{I}]$ and $\mathbf{Q}[\mathcal{I},:]$, as illustrated in Figure~\ref{fig:intro-draft} (left). Compared to the original multiplication procedure, AMM reduces the computational complexity to $|\mathcal{I}|/n$ times by instead multiplying $\mathbf{P}[:,\mathcal{I}]$ with $\mathbf{Q}[\mathcal{I},:]$, trading computational accuracy for computational efficiency. To identify the important rows or columns, we compute the importance scores $\alpha_i=\|\mathbf{P}[:,i]\mathbf{Q}[i,:]\|_F$ every $\tau$ iterations and select the top ones in the following $\tau$ times of AMM operation. 

Unfortunately, the computational error in the activations' gradients can further propagate to previous layers. Under such a long-lasting effect of inaccurate activation gradients, the  accumulated computational error can lead to poor gradient estimations, which severely harms the optimization procedure and degrades the final model performance. To resolve this issue, we develop a double-LoRA technique that significantly reduces the relative error induced to the activation gradients by AMM. Specifically, double-LoRA splits the frozen dense weight matrix into two parts, where the first part applies AMM to save computation, the second part is a frozen LoRA adapter which is computation-efficient without using AMM. In order to reduce the AMM-induced error, we expect the first part to include as little information as possible. Consequently, we initially conduct SVD of the parameter matrix and identify the first part according to the smallest singular values and corresponding vectors.
Combining the above techniques, we propose \textub{C}omputation-\textub{E}fficient \textub{LoRA} (\celora), which is more computation-efficient and equally memory-efficient compared with LoRA.

We evaluate the proposed \celora algorithm both theoretically and empirically. In theory, we prove that \celora with momentum SGD converges at a rate of $\mathcal{O}(1/\sqrt{T})$, the same order of LoRA's convergence rate. Empirically, we validate that \celora can converge at a comparable precision as standard LoRA, with slightly reduced memory consumption and a 3.39$\times$ acceleration in computation. To our knowledge, \celora is the \textit{first} algorithm that accelerates LoRA without sacrificing memory-efficiency or leading to notable performance degradation.

The main contributions of this paper are as follows:

\begin{itemize}[topsep=5pt, leftmargin=1em]

    \vspace{-0.5em}
    \item We propose a novel algorithm \celora, which is more computation-efficient and equally memory-efficient compared with standard LoRA.

    \item We theoretically prove that \celora converges at a rate of $\mathcal{O}(1/\sqrt{T})$, which is the same order of standard LoRA's convergence rate.

    \item We experimentally validate that \celora has a 3.39$\times$ acceleration compared with LoRA without sacrificing memory-efficiency or leading to notable training performance degradation.
\end{itemize}

\section{Preliminaries}\label{sec:preliminaries}


\subsection{LoRA Algorithm}
In order to fine-tune language models memory-efficiently, LoRA applies a low-rank adapter to each linear layer in the model. Specifically, let $\mathbf{y}=\mathbf{W}\mathbf{x}$ represent a linear layer with $\mathbf{y}\in\mathbb{R}^{m\times b}$, $\mathbf{W}\in\mathbb{R}^{m\times n}$ and $\mathbf{x}\in\mathbb{R}^{n\times b}$, where $m,n$ represent the output and input dimension, respectively, and $b$ represents the batch-size. The LoRA adapter is given by $\mathbf{W}=\mathbf{W}_0+\mathbf{B}\mathbf{A}$ with $\mathbf{W}_0\in\mathbb{R}^{m\times n}$ fixed as the pre-trained weights, $\mathbf{B}\in\mathbb{R}^{m\times r}$ and $\mathbf{A}\in\mathbb{R}^{r\times n}$ trainable.
\begin{align*}
    \mathbf{y}=&\ \mathbf{W}\mathbf{x},&\quad\text{(Original)}\\
    \mathbf{y}=&\ (\mathbf{W}_0+\mathbf{B}\mathbf{A})\mathbf{x}.&\quad\text{(LoRA)}
\end{align*}
When $r\ll\min\{m,n\}$, the number of trainable parameters in LoRA, $(m+n)r$, is much fewer than that in full fine-tuning, \ie, $mn$, which significantly reduces the memory consumption of the optimizer states. 

\subsection{Computational Bottleneck}
While applying LoRA can significantly reduce the memory cost for fine-tuning large language models, the computational cost for computing the gradient via back propagation is not sufficiently reduced. Specifically, let $\mathbf{G}_{\mathbf{\theta}}$ denote the stochastic gradient of $\mathbf{\theta}$ calculated by back propagation, for linear layer $\mathbf{y}=\mathbf{W}\mathbf{x}$ we compute:
\begin{align}
\mathbf{G}_\mathbf{W}=&\ \mathbf{G}_\mathbf{y}\cdot\mathbf{x}^\top,\label{eq:W1}\\
\mathbf{G}_\mathbf{x}=&\ \mathbf{W}^\top\cdot \mathbf{G}_\mathbf{y},\label{eq:A1}
\end{align}
Both \eqref{eq:W1} and \eqref{eq:A1} require $2bmn$ FLOPs of computation. Accordingly, in LoRA we compute:
\begin{align}
    \mathbf{G}_\mathbf{B}=&\ \mathbf{G}_\mathbf{y}\cdot\mathbf{z}^\top,\label{eq:W'1}\\
    \mathbf{G}_\mathbf{z}=&\ \mathbf{B}^\top\cdot \mathbf{G}_\mathbf{y},\label{eq:A'1}\\
    \mathbf{G}_\mathbf{A}=&\ \mathbf{G}_\mathbf{z}\cdot\mathbf{x}^\top,\label{eq:W'2}\\
\mathbf{G}_{\mathbf{x},1}=&\ \mathbf{A}^\top\cdot \mathbf{G}_\mathbf{z},\label{eq:A'2}\\
    \mathbf{G}_{\mathbf{x},2}=&\ \mathbf{W}_0^\top\cdot \mathbf{G}_\mathbf{y},\label{eq:A'3}\\
    \mathbf{G}_\mathbf{x}=&\ \mathbf{G}_{\mathbf{x},1}+\mathbf{G}_{\mathbf{x},2},\label{eq:A'4}
\end{align}
where $\mathbf{z}=\mathbf{A}\mathbf{x}\in\mathbb{R}^{r\times b}$ represents LoRA's additional activation. \eqref{eq:W'1}\eqref{eq:A'1}\eqref{eq:W'2}\eqref{eq:A'2}\eqref{eq:A'3}\eqref{eq:A'4} require $2brm$, $2brm$, $2brn$, $2brn$, $2bmn$, and $bn$ FLOPs of computation, respectively, resulting in $4br(m+n)+2bmn+bn$ FLOPs in total. When $r\ll\min\{m,n\}$, the computational cost in LoRA's back propagation is roughly $2bmn$, half of the computation in the original approach \eqref{eq:W1}\eqref{eq:A1}. The computational bottleneck lies in the dense matrix multiplication step in \eqref{eq:A'3}, which alone requires $2bmn$ FLOPs of computation.

\section{\celora: Computation-Efficient LoRA}


\subsection{Approximated Matrix Multiplication (AMM)}
Consider matrix multiplication $\mathbf{T}=\mathbf{P}\mathbf{Q}$, where $\mathbf{T}\in\mathbb{R}^{m\times k}$, $\mathbf{P}\in\mathbb{R}^{m\times n}$ and $\mathbf{Q}\in\mathbb{R}^{n\times k}$.
Let $\mathbf{p}_1,\mathbf{p}_2,\cdots,\mathbf{p}_n$ denote the column vectors of matrix $\mathbf{P}$, and $\mathbf{q}_1,\mathbf{q}_2,\cdots,\mathbf{q}_n$ denote the column vectors of matrix $\mathbf{Q}^\top$. 
We can rewrite the matrix multiplication into:
\begin{align*}
    \mathbf{T}=\sum_{i=1}^n\mathbf{p}_i\mathbf{q}_i^\top.
\end{align*}
To estimate the product $\mathbf{T}$ computation-efficiently, we may assume the matrices $\mathbf{P}$ and $\mathbf{Q}$ enjoy some kinds of structured sparsity, such that a few $(\mathbf{p}_i\mathbf{q}_i^\top)'s$ contribute to most of the result $\sum_{i=1}^n\mathbf{p}_i\mathbf{q}_i^\top$, in which case we could estimate $\mathbf{T}$ by computing the most important parts only. Specifically, we identify $s$ most important indices $1\le i_1<\cdots<i_s\le n$, and the AMM estimate of $\mathbf{T}$ is given by:
\begin{align*}
    \hat{\mathbf{T}}=\sum_{j=1}^s{\mathbf{p}_{i_j}\mathbf{q}_{i_j}^\top}=\hat{\mathbf{P}}\hat{\mathbf{Q}},
\end{align*}
where $\hat{\mathbf{P}}$ and $\hat{\mathbf{Q}}^\top$ collect column vectors $\{\mathbf{p}_{i_j}\}_{j=1}^s$ and $\{\mathbf{q}_{i_j}\}_{j=1}^s$, respectively.

The efficiency of AMM is concerned with the number of selected indices $s$, or the structured sparsity  $p:=s/n\in(0,1]$. Replacing the dense matrix multiplication $\mathbf{T}=\mathbf{P}\mathbf{Q}$ by AMM estimate  $\hat{\mathbf{T}}=\hat{\mathbf{P}}\hat{\mathbf{Q}}$, the computational complexity is reduced from $2mnk$ to $2msk=p\cdot (2mnk)$.
Hereafter, we use $\mathcal{C}_p(\mathbf{P}\cdot\mathbf{Q})$ to denote the AMM estimate of matrix multiplication $\mathbf{P}\cdot\mathbf{Q}$ with structured sparsity $p$.

An important question is how to select the indices $\mathcal{I}=\{i_1,i_2,\cdots,i_s\}$ properly. A previous research \citep{drineas2006fast} has studied a random sampling strategy, which does not work well in our experiments. Based on the above intuition, we define the importance score $\alpha_i$ of index $i$ by the Frobenius norm $\|\mathbf{p}_i\mathbf{q}_i^\top\|_F$ and attempt to select the indices with highest scores. However, as calculating $\{\alpha_i\}_{i=1}^k$ requires the same amount of computation as that of conducting the original matrix multiplication, we cannot determine $\mathcal{I}$ based on the calculation results of $\{\alpha_i\}_{i=1}^k$ in every iteration. We use historical information to mitigate this issue. Specifically, the matrices $\mathbf{P},\mathbf{Q}$ we multiply by AMM should be variables that live along the whole optimization process, and $\mathbf{P}^t,\mathbf{Q}^t$ are multiplied at every iteration $t$. The corresponding $\mathcal{I}^t$ is only re-selected according to the top-$s$ importance scores every $\tau$ iterations and is reused in intermediate ones.

To reduce the computational bottleneck in LoRA's backward propagation, we apply AMM to step \eqref{eq:A'3} and get:
\begin{align}
\mathbf{G}_{\mathbf{x},2}=&\ \mathcal{C}_p(\mathbf{W}_0^\top\cdot \mathbf{G}_\mathbf{y}).\label{eq:A''3}
\end{align}

\subsection{Double-LoRA Mechanism }
Although computation-efficient, AMM will induce errors to $g_x$, the gradient with respect to the activations. 
These errors propagate backward through the network, potentially compounding as they traverse previous layers.
If the magnitude of these errors is not properly controlled, the accuracy of the parameter gradients can be significantly degraded.
To mitigate this issue, we propose a double-LoRA mechanism to alleviate the error induced by the AMM operation in each layer. Intuitively, we wish the objective matrix multiplication result we estimate by AMM has as little contribution to the activation gradient as possible. This drives us to further separate the frozen matrix $\mathbf{W}_0$ into two parts: a low-rank part inheriting computational efficiency without AMM, and a residual part with a relatively small magnitude.

\begin{figure}[!t]
    \centering
    \includegraphics[width=\linewidth]{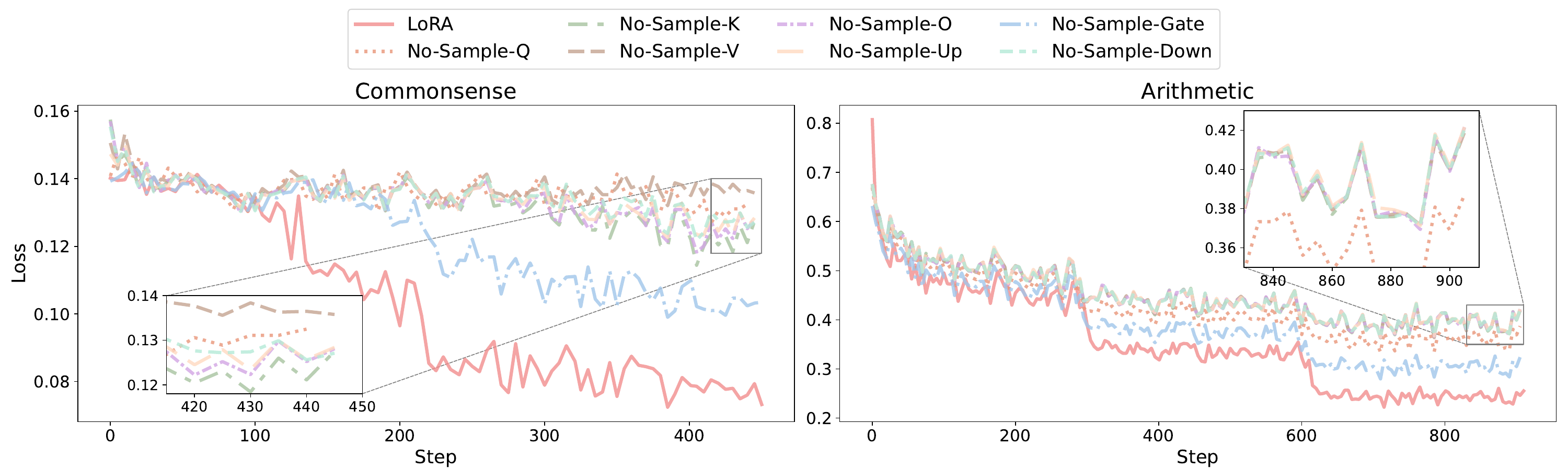}
    \vspace{-1em}
    \caption{\small
    Layer-wise Sensitivity Analysis of LLaMA3.2-1B. 
    }
    \label{fig:layerwise}
    \vspace{-1em}    
\end{figure}

Specifically, we initially compute the SVD of $\mathbf{W}_0$, yielding
\begin{align*}
    \mathbf{W}_0=\mathbf{U}\mathbf{\Sigma} \mathbf{V}^\top
\end{align*}
Next, We collect the principal low-rank component $\mathbf{B}_0=\mathbf{U}[:r]\mathbf{\Sigma}^{1/2}$, $\mathbf{A}_0=\mathbf{\Sigma}^{1/2}(\mathbf{V}[:r])^\top$, and the residual $\mathbf{W}_{s}=\mathbf{W}_0-\mathbf{B}_0\mathbf{A}_0$. By separating $\mathbf{W}_0$ to $\mathbf{W}_s+\mathbf{B}_0\mathbf{A}_0$, we split the matrix to two parts. The first part $\mathbf{W}_s$ is believed to have better structured sparsity and is more compatible to AMM. The second low-rank part $\mathbf{B}_0\mathbf{A}_0$ is computation-efficient just like the trainable LoRA adapter $\mathbf{B}\mathbf{A}$. Combining AMM with double-LoRA, \eqref{eq:A''3} is further replaced by
\begin{align}
\mathbf{G}_{\mathbf{x},2}=&\ \mathcal{C}_p(\mathbf{W}_s^\top\cdot \mathbf{G}_\mathbf{y})+\mathbf{A}_0^\top(\mathbf{B}_0^\top \mathbf{G}_\mathbf{y}).\label{eq:A'''3}
\end{align}

\subsection{Layer-wise Adaptive Sparsity}

It is natural to apply more aggressive sparsity to layers that are relatively robust to computational errors, while using more conservative sparsity for those that are more sensitive.
Inspired by \cite{hu2025accelerating,ma2024first,jaiswal2024galore,zeng2024lsaq,malinovskii2024pushing,zhang2024q,liu2024training}, we adopt a layer-wise adaptive sparsity strategy for \celora.

To determine which layers are more sensitive to varying sparsity levels, we conduct experiments on two small fine-tuning datasets: the Commonsense 14K dataset and the Math 7K dataset~\cite{hu2023llm}.
In these experiments, we fix LoRA's rank to 32, and set both \celora's trainable LoRA rank and its frozen Double-LoRA rank to 28. 
For each \celora configuration, we vary the sparsity level of one layer type, while setting the sparsity of all remaining layer types to $p=0.3$.
As shown in Figure~\ref{fig:layerwise}, the \texttt{Gate} layers are essential for preventing error propagation. 
In addition, the \texttt{Q} and \texttt{K} layers have a strong impact on arithmetic and commonsense reasoning tasks, respectively. 
Based on these findings, we disable sparsity for the \texttt{Q}, \texttt{K}, and \texttt{Gate} layers. 
For the remaining MHA layers, we use $p=0.55$, and for the last two layers in the FFN, we set $p=0.65$ throughout our experiments.

\subsection{Algorithm}
\begin{algorithm}[t]
\caption{\celora}\label{alg:CeLoRA}
\footnotesize
\begin{algorithmic}[1]

\Statex{\textbf{Input:} Frozen layer weight $\mathbf{W}_{\ell}\in\mathbb{R}^{m_{\ell}\times n_{\ell}}$, sparsity level $p_\ell$, double-LoRA rank $r_{0,\ell}$, indices recomputing period $\tau$, Top-K indices $\mathcal{I}_\ell=$ empty, optimizer $\rho$.
}

\Statex\rule{\linewidth}{0.4pt}
\Statex \textbf{Initialize Double-LoRA}
    \State \quad \textbf{for} Layer $\ell=1,2,\cdots,L$ \textbf{do}
    \State \qquad Conducting SVD on frozen weight matrix
    \Statex \qquad $\mathbf{W}_{0,\ell}=\mathbf{U}_\ell\mathbf{\Sigma}_\ell \mathbf{V}_\ell^\top$;
    \State \qquad $\mathbf{A}_{0,\ell},\ \mathbf{B}_{0,\ell} \gets \sqrt{\mathbf{\Sigma}_\ell}{\mathbf{V}_\ell^\top}_{[:r_0,]},\ {\mathbf{U}_{\ell}}_{[,:r_0]}\sqrt{\mathbf{\Sigma}_\ell}$ ;\Comment{Stored in layer's buffer.}
    \State \quad \textbf{end for}
\Statex \rule{\linewidth}{0.4pt}

\State \textbf{for} {\textbf{\celora Training Step $t=0,1,\cdots,T-1$}} \textbf{do}
\vspace{0.5mm}
    \State \quad \textbf{for} Layer $\ell=1,2,\cdots,L$ \textbf{do}\Comment{\textbf{Forward}}
        \State \quad\quad $\mathbf{z}_\ell\gets\mathbf{A}_\ell\mathbf{x}_\ell$;
        \State \quad\quad $\mathbf{y}_\ell \gets \mathbf{W}_{0,\ell}\mathbf{x}_\ell+\mathbf{B}_\ell\mathbf{z}_\ell$;
    \State\quad\textbf{end for}
\vspace{0.5mm}
    \State \quad \textbf{for} Layer $\ell=L,L-1,\cdots,1$ \textbf{do}\Comment{\textbf{Backward}}
        \State \quad\quad  $\mathbf{W}_{s,\ell} \gets \mathbf{W}_{0,\ell} - \mathbf{B}_{0,\ell}\mathbf{A}_{0,\ell}$;
        \State \quad\quad \textbf{if} {$\tau\mid t$ \textbf{or} $\mathcal{I}_\ell$ is empty} \textbf{then}
            
            \State \quad\qquad $\alpha_{i,\ell} \gets\left\|{\mathbf{W}_{s,\ell}^\top}_{[:,i]} {\mathbf{G_{y_\ell}}}_{[i,:]}\right\|_F$, \ $\forall i \in \left\{1,\dots,m_\ell\right\}$
            \State \quad\qquad {Select $\left\{i_{1,\ell},\cdots,i_{\text{K}_\ell,\ell}\right\}$ with largest $\alpha_{i,\ell}$'s;}
            
            \State \quad\qquad $\mathcal{I}_\ell = \left\{i_{1,\ell},\cdots,i_{\text{K}_\ell,\ell}\right\}$; \Comment{Here K$_\ell= \lceil m_\ell p_\ell \rceil$}
        \State \quad\quad\textbf{end if}
        \State \quad\quad $\mathbf{G}_{\mathbf{B}_\ell}\gets\mathbf{G}_{\mathbf{y}_\ell}\mathbf{z}_\ell^\top$;
        \State \quad\quad $\mathbf{G}_{\mathbf{z}_\ell}\gets\mathbf{B}_{\ell}^\top\mathbf{G}_{\mathbf{y}_{\ell}}$;
        \State \quad\quad
        $\mathbf{G}_{\mathbf{A}_\ell}\gets\mathbf{G}_{\mathbf{z}_\ell}\mathbf{x}_\ell^\top$;
        \State \quad\quad $\mathbf{G}_{\mathbf{x}_\ell} \gets {\mathbf{W}_{s,\ell}^\top}_{[,\mathcal{I}]} {\mathbf{G}_{\mathbf{y}_\ell}}_{[\mathcal{I},]} + \mathbf{A}_{0,\ell}^\top(\mathbf{B}_{0,\ell}^\top \mathbf{G}_{\mathbf{y}_\ell})+\mathbf{A}_\ell^\top\mathbf{G}_{\mathbf{z}_\ell}$;
\State\quad\textbf{end for}
\State Use optimizer $\rho$ to update $\{\mathbf{A}_\ell,\mathbf{B}_\ell\}_{\ell=1}^L$ according to $\{\mathbf{G}_{\mathbf{A}_\ell},\mathbf{G}_{\mathbf{B}_\ell}\}_{\ell=1}^L$;
\State\textbf{end for}
\end{algorithmic}
\end{algorithm}

Overall, \celora integrates AMM, double-LoRA, and layer-wise adaptivity, as outlined in Algorithm~\ref{alg:CeLoRA}.
During model initialization, we replace all frozen linear layers with \celora and apply the double-LoRA technique to the weight matrix $\mathbf{W}_0$, resulting in low-rank components $\mathbf{A}_0$ and $\mathbf{B}_0$ (lines 2–3). 
For each training step $t$, the forward pass of a \celora linear layer behaves the same as the original frozen linear layer (lines 8).
In the backward pass, \celora first computes the residual weight matrix by subtracting the low-rank components from the original weight matrix (line 11). 
Next, if the current step $t$ is a multiple of $\tau$ or if the indices are empty (\eg, at the start of training), the top-K indices are updated (lines 12–15). 
Finally, \celora uses AMM to compute activation gradient $\mathbf{G}_\mathbf{x_\ell}$ (line 20).

\begin{table*}[!b]
    \centering
    \caption{Computation and memory analysis for a single linear layer.}
    \label{tab:complexity}
    \vspace{0.5em}
    \begin{adjustbox}{max width=\textwidth}
    \begin{tabular}{cccc}
    \toprule
         \textbf{Method} & Standard AdamW & LoRA & \celora\\
    \midrule
    \multirow{2}{*}{\textbf{Memory Usage}} & \multirow{2}{*}{$10mn+2bm$} & $2mn+10r(m+n)$ & $2mn+2r_0(m+n)$\\
    & &$+2b(m+r)$ & {$+10r(m+n)+2b(m+r)$}\\
    \midrule
    \textbf{Forward Computation} & $2bmn$ & $2bmn+2br(m+n)$ & $2bmn+2br(m+n)$\\
    \midrule
    \multirow{2}{*}{\textbf{Backward Computation}} & \multirow{2}{*}{$4bmn$} & \multirow{2}{*}{$2bmn+4br(m+n)$} & $(2pb+1)mn$\\
    & & & $+2(r_0 + br_0 + 2br)(m+n)$\\
    \bottomrule
    \end{tabular}
    \end{adjustbox}
\end{table*}
\subsection{Complexity Analysis}
To better illustrate the computational efficiency of \celora, we theoretically compare the computational and memory complexity of \celora with LoRA and standard AdamW fine-tuning. Consider linear layer $\mathbf{y}=\mathbf{W}\mathbf{x}$ with $\mathbf{W}\in\mathbb{R}^{m\times n}$, trained with LoRA rank $r$, double-LoRA rank $r_0$, structured sparsity $p$ and batch size $b$ using BF16 precision. As illustrated in Table~\ref{tab:complexity}, \celora can achieve a memory usage similar to LoRA by applying slightly smaller $r_0$ and $r$, while significantly reduce the backward computation by applying a relatively small $p$ when $b\gg 1$ and $r\ll\min\{m,n\}$. 
When combined with low-precision training, the influence of double-LoRA can be further reduced, as the frozen low-rank parameters do not require high-precision weight copies or gradient accumulators.

\section{Convergence Analysis}

We first present the assumptions under which we prove \celora's convergence properties. 

\begin{assumption}[Lower Boundedness]\label{asp:proper}
    The loss function $f:\mathbb{R}^{d}\rightarrow\mathbb{R}$ satisfies $\inf_{\mathbf{x}\in\mathbb{R}^d}f(\mathbf{x})>-\infty$.
\end{assumption}
\begin{assumption}[$L$-Smoothness]\label{asp:smoothness}
    The loss function $f$ is $L$-smooth, \ie, it holds for any $\mathbf{x},\mathbf{y}\in\mathbb{R}^{d}$ that
    \begin{align*}
        \|\nabla f(\mathbf{x})-\nabla f(\mathbf{y})\|_2\le L\|\mathbf{x}-\mathbf{y}\|_2.
    \end{align*}
\end{assumption}
\begin{assumption}[Stochastic Gradient]\label{asp:stochastic}
    We assume the stochastic gradient oracle satisfies
    \begin{align}
        &\mathbb{E}[\nabla F(\mathbf{x}^t;\xi^t)]=\nabla f(\mathbf{x}^t);\\
        &\mathbb{E}[\|\nabla F(\mathbf{x}^t;\xi^t)-\nabla f(\mathbf{x}^t)\|^2]\le\sigma^2,
    \end{align}
    for some $\sigma>0$.
\end{assumption}

Assumptions \ref{asp:proper}-\ref{asp:stochastic} are standard assumptions commonly used in stochastic optimization. 

\begin{assumption}[Gradient Error]\label{asp:contractive}
    Let $\mathbf{g}^t$ and $\hat{\mathbf{g}}^t$ denote the original stochastic gradient $\nabla F(\mathbf{x}^t,\xi^t)$ and its estimation by \celora, it holds that
    \begin{align}
        \|\hat{\mathbf{g}}^t-\mathbf{g}^t\|^2\le(1-\delta)\|\mathbf{g}^t\|^2,\label{eq:asp-cgk}
    \end{align}
    and 
    \begin{align}
        \|\mathbb{E}_{\xi^t\sim\mathcal{D}}[\hat{\mathbf{g}}^t]-\nabla f(\mathbf{x}^t)\|^2\le(1-\delta)\|\nabla f(\mathbf{x}^t)\|^2,\label{eq:asp-ecgk}
    \end{align}
    for some $\delta\in(0,1].$
\end{assumption}

Assumption \ref{asp:contractive} illustrates the property of stochastic gradients in \celora. Though not standard, this assumption can be empirically justified by our experimental results.

\begin{figure*}[ht]
    \centering
        \includegraphics[width=\textwidth]{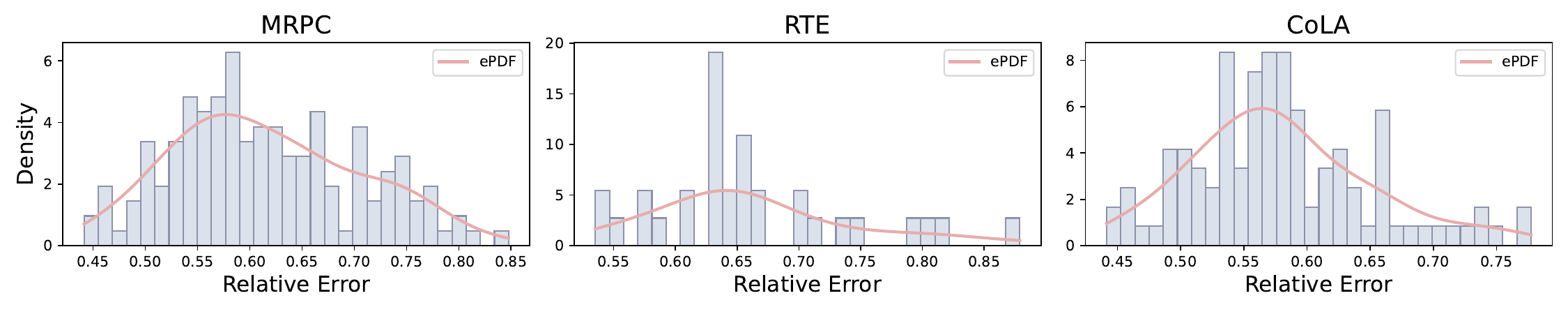}
    \vspace{-4mm}
    \caption{Empirical validation of \eqref{eq:asp-cgk} on MRPC (left), RTE (middle) and CoLA (right).}
    \label{fig:cgk}
\end{figure*}

\begin{figure*}[ht]
    \centering
        \includegraphics[width=\textwidth]{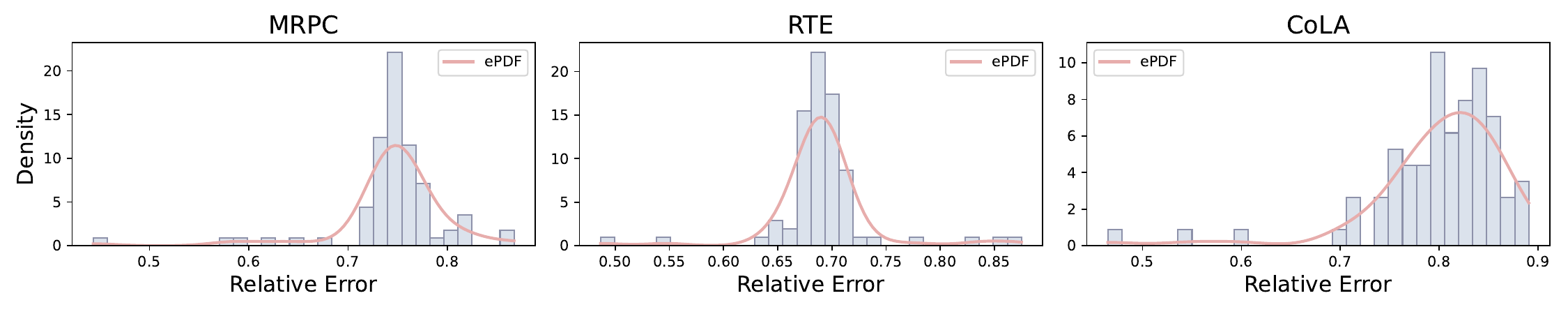}
    \vspace{-4mm}
    \caption{Empirical validation of \eqref{eq:asp-ecgk} on MRPC (left), RTE (middle) and CoLA (right).}
    \label{fig:ecgk}
\end{figure*}

\textbf{Empirical Justification of Assumption \ref{asp:contractive}.} To justify \eqref{eq:asp-cgk}, we conduct experiments on language model fine-tuning tasks on google/gemma-2b~\cite{gemmamodelcard} model using CoLA~\cite{warstadt-etal-2019-neural}, RTE and MRPC~\cite{dolan-brockett-2005-automatically} datasets, three tasks in the GLUE benchmark~\cite{wang2019gluemultitaskbenchmarkanalysis}. In these experiments, we use AdamW with a learning rate of \texttt{1e-5} to train for 1 epoch per task and calculate the relative error $\|\hat{\mathbf{g}}^t-\mathbf{g}^t\|^2/\|\mathbf{g}^t\|^2$ every 10 steps, as illustrated in Figure~\ref{fig:cgk}, where all relative errors are below 0.9. To justify \eqref{eq:asp-ecgk}, we conduct experiments on the same model and datasets, where we alternatively calculate one iteration of full gradient AdamW and one epoch of random gradient AdamW, each with a learning rate of \texttt{1e-5} for a total of 80 cycles. We apply a rank of 64 for both LoRA and double-LoRA in \celora, and apply a structured sparsity of $p_{\mathrm{FFN}}=0.9$ and $p_{\mathrm{MHA}}=0.4$. We calculate the relative error $\|\mathbb{E}_{\xi^t\sim\mathcal{D}}[\hat{\mathbf{g}}^t]-\nabla f(\mathbf{x}^t)\|^2/\|\nabla f(\mathbf{x}^t)\|^2$ for every full-gradient step, as illustrated in Figure~\ref{fig:ecgk}, where all relative errors are below 0.9.

We now propose the convergence results of \celora using  the momentum SGD optimizer with the following update:
\begin{align*}
\mathbf{m}^t=&\ (1-\beta_1)\mathbf{m}^{t-1}+\beta_1\hat{\mathbf{g}}^t,\\
\mathbf{x}^{t+1}=&\ \mathbf{x}^t-\eta \mathbf{m}^t,
\end{align*}
\begin{theorem}\label{thm:celora}
    Under Assumptions \ref{asp:proper} - \ref{asp:contractive}, if $\beta_1\in\left(0,\frac{\delta}{24-12\delta}\right)$ and $\eta\le\min\left\{\frac{L}{2}, \frac{\beta_1}{L}\cdot\sqrt{\frac{\delta}{8}}\right\}$, \celora with momentum SGD converges as
    \begin{align*}
        &\frac{1}{T+1}\sum_{t=0}^T\mathbb{E}[\|\nabla f(\mathbf{x}^t)\|_2^2]
        \le\frac{4[f(\mathbf{x}^0)-\inf_{\mathbf{x}}f(\mathbf{x})]}{\delta\eta(T+1)}+\frac{4\|\mathbf{m}^0-\nabla f(\mathbf{x}^0)\|_2^2]}{\delta\beta_1(T+1)}+\frac{12\beta_1\sigma^2}{\delta}. \nonumber
    \end{align*}
\end{theorem}
\begin{corollary}
    Under Assumptions \ref{asp:proper}-\ref{asp:contractive}, if we choose $\beta_1=\left(\frac{24}{\delta}+\sigma\sqrt{\frac{\delta^{1/2}\left(T+1\right)}{L\Delta}}\right)^{-1}$, $\eta=\left(2L+ \frac{2^{3/2}L}{\delta^{1/2}\beta_1}\right)^{-1}$, \celora with momentum SGD converges as
    \begin{align}
        &\frac{1}{T+1}\sum_{t=0}^{T}\mathbb{E}[\|\nabla f(\mathbf{x}^t)\|_2^2]
        =\mathcal{O}\left(\frac{L\Delta}{\delta^{5/2}(T+1)}+\sqrt{\frac{L\Delta\sigma^2}{\delta^{5/2}(T+1)}}\right),\nonumber
    \end{align}
    where $\Delta:=f(\mathbf{x}^0)-\inf_{\mathbf{x}}f(\mathbf{x})+(\delta/L)\cdot\|\mathbf{m}^0-\nabla f(\mathbf{x}^0)\|_2^2$.
\end{corollary}

Detailed proofs are deferred to Appendix \ref{app:proof}.

\section{Experiments}

In this section, we present a comprehensive set of experiments to evaluate the convergence performance and computational efficiency of \celora, and compare it against the baseline method.

\subsection{Experimental Setup}

\textbf{Datasets.}
We follow the benchmark design outlined in \cite{hu2023llm} and evaluate \celora on two popular reasoning benchmarks:

\begin{itemize}[topsep=5pt, leftmargin=1em]
\vspace{-0.5em}
\item \textbf{Commonsense Reasoning}: This dataset includes eight tasks: BoolQ \cite{clark-etal-2019-boolq}, PIQA \cite{bisk2020piqa}, SocialQA \cite{sap2019socialiqa}, HellaSwag \cite{zellers2019hellaswag}, WinoGrande \cite{sakaguchi2021winogrande}, ARC-challenge \cite{clark2018think}, ARC-easy \cite{clark2018think}, and OpenbookQA \cite{mihaylov2018can}. In our experiments, we fine-tune all models using the Commonsense 170K dataset \cite{hu2023llm}, which is constructed by combining the training sets from these eight tasks.

\vspace{-0.5em}
\item \textbf{Arithmetic Reasoning}:
This benchmark consists of seven subsets: MultiArith \cite{roy2016solving}, GSM8K \cite{cobbe2021training}, AddSub \cite{hosseini2014learning}, AQuA \cite{ling-etal-2017-program}, SingleEq \cite{koncel-kedziorski-etal-2015-parsing}, SVAMP \cite{patel-etal-2021-nlp} and MAWPS \cite{koncel-kedziorski-etal-2016-mawps}.
We fine-tune the models on the Math 10k dataset \cite{hu2023llm}, which includes training data from GSM8K, MAWPS, and AQuA, augmented by language models with chain-of-thought reasoning steps.
\vspace{-0.5em}
\end{itemize}

\textbf{Fine-tuned models and hyper-parameters.} We fine-tune LLaMA-2-7B, LLaMA-2-13B \cite{touvron2023llama2openfoundation}, and LLaMA-3.1-8B \cite{llama3modelcard} using both \celora and LoRA.
The adapter is applied to all linear layers in each transformer block, including \texttt{Q}, \texttt{K}, \texttt{V}, \texttt{O}, \texttt{Up}, \texttt{Gate}, and \texttt{Down}.
Unless specified otherwise, all \celora experiments replace the frozen \texttt{V}, \texttt{O}, \texttt{Up}, and \texttt{Down} layers with \celora layers.
The sparsity levels are set as follows: $p_\text{V} = p_\text{O} = 0.55$ and $p_\text{Up} = p_\text{Down} = 0.65$. 
For consistency, the same set of hyperparameters is applied across both methods for each model size.
All experiments are conducted using the BF16 format to optimize memory usage.

\begin{table*}[!b]
    \caption{Comparison among eight commonsense reasoning tasks for the LLaMA2-7B/13B, LLaMA3.1-8B models}
    \vspace{0.5em}
    \label{table:common_accuracy}
    \centering
    \begin{adjustbox}{max width=\textwidth}

    \begin{tabular}{l l c c c c c c c c c c}
        \toprule
        \textbf{Model} & \textbf{Method} & \textbf{Rank} & \textbf{BoolQ} & \textbf{PIQA} & \textbf{SIQA} & \textbf{HellaSwag} & \textbf{Wino} & \textbf{ARC-e} & \textbf{ARC-c} & \textbf{OBQA} & \textbf{Avg. $\uparrow$} \\
        \midrule
        \multirow{4}{*}{LLaMA2-7B} 
            & LoRA & 16 & 71.99 & 84.49 & 81.73 & 94.45 & 85.95 & 87.63 & 73.21 & 83.80 & 82.91 \\
            & \cellcolor{skyblue}\celora & \cellcolor{skyblue}14 & \cellcolor{skyblue}70.24 & \cellcolor{skyblue}82.59 & \cellcolor{skyblue}79.27 & \cellcolor{skyblue}93.17 & \cellcolor{skyblue}82.72 & \cellcolor{skyblue}85.56 & \cellcolor{skyblue}70.65 & \cellcolor{skyblue}79.80 & \cellcolor{skyblue}80.50 \\
            & LoRA & 64 & 72.26 & 84.88 & 82.70 & 94.97 & 86.42 & 88.55 & 74.74 & 86.40 & 83.87 \\
            & \cellcolor{skyblue}\celora & \cellcolor{skyblue}56 & \cellcolor{skyblue}71.68 & \cellcolor{skyblue}85.20 & \cellcolor{skyblue}82.09 & \cellcolor{skyblue}94.61 & \cellcolor{skyblue}83.98 & \cellcolor{skyblue}87.29 & \cellcolor{skyblue}73.29 & \cellcolor{skyblue}84.20 & \cellcolor{skyblue}82.79 \\
        \midrule
        \multirow{4}{*}{LLaMA2-13B}
            & LoRA & 16 & 75.32 & 88.03 & 83.21 & 81.14 & 92.34 & 88.20 & 96.08 & 88.71 & 86.63 \\
            & \cellcolor{skyblue}\celora & \cellcolor{skyblue}14 & \cellcolor{skyblue}73.21 & \cellcolor{skyblue}86.62 & \cellcolor{skyblue}82.14 & \cellcolor{skyblue}94.65 & \cellcolor{skyblue}86.27 & \cellcolor{skyblue}90.15 & \cellcolor{skyblue}77.22 & \cellcolor{skyblue}84.80 & \cellcolor{skyblue}84.38 \\
            & LoRA & 64 & 75.72 & 88.85 & 84.39 & 96.34 & 88.71 & 92.42 & 81.83 & 89.60 & 87.23 \\
            & \cellcolor{skyblue}\celora & \cellcolor{skyblue}56 & \cellcolor{skyblue}74.01 & \cellcolor{skyblue}86.51 & \cellcolor{skyblue}83.11 & \cellcolor{skyblue}92.74 & \cellcolor{skyblue}87.92 & \cellcolor{skyblue}91.37 & \cellcolor{skyblue}79.61 & \cellcolor{skyblue}85.40 & \cellcolor{skyblue}85.08 \\
        \midrule
        \multirow{4}{*}{LLaMA3-8B} 
            & LoRA & 16 & 75.84 & 90.86 & 83.52 & 96.93 & 89.90 & 94.07 & 84.47 & 88.8 & 88.05 \\
            & \cellcolor{skyblue}\celora & \cellcolor{skyblue}14 & \cellcolor{skyblue}72.08 & \cellcolor{skyblue}89.72 & \cellcolor{skyblue}82.65 & \cellcolor{skyblue}96.24 & \cellcolor{skyblue}88.32 & \cellcolor{skyblue}93.35 & \cellcolor{skyblue}83.36 & \cellcolor{skyblue}87.60 & \cellcolor{skyblue}86.66 \\
            & LoRA & 64 & 75.63 & 90.21 & 83.32 & 96.38 & 88.95 & 93.39 & 84.04 & 89.20 & 87.64 \\
            & \cellcolor{skyblue}\celora & \cellcolor{skyblue}56 & \cellcolor{skyblue}73.36 & \cellcolor{skyblue}89.66 & \cellcolor{skyblue}82.40 & \cellcolor{skyblue}95.76 & \cellcolor{skyblue}86.42 & \cellcolor{skyblue}93.14 & \cellcolor{skyblue}82.68 & \cellcolor{skyblue}87.60 & \cellcolor{skyblue}86.38 \\
        \bottomrule

    \end{tabular}
    \end{adjustbox}
\end{table*}

\subsection{Statistical Efficiency of \celora}

\begin{figure*}[!t]
  \centering
    \includegraphics[width=\linewidth]{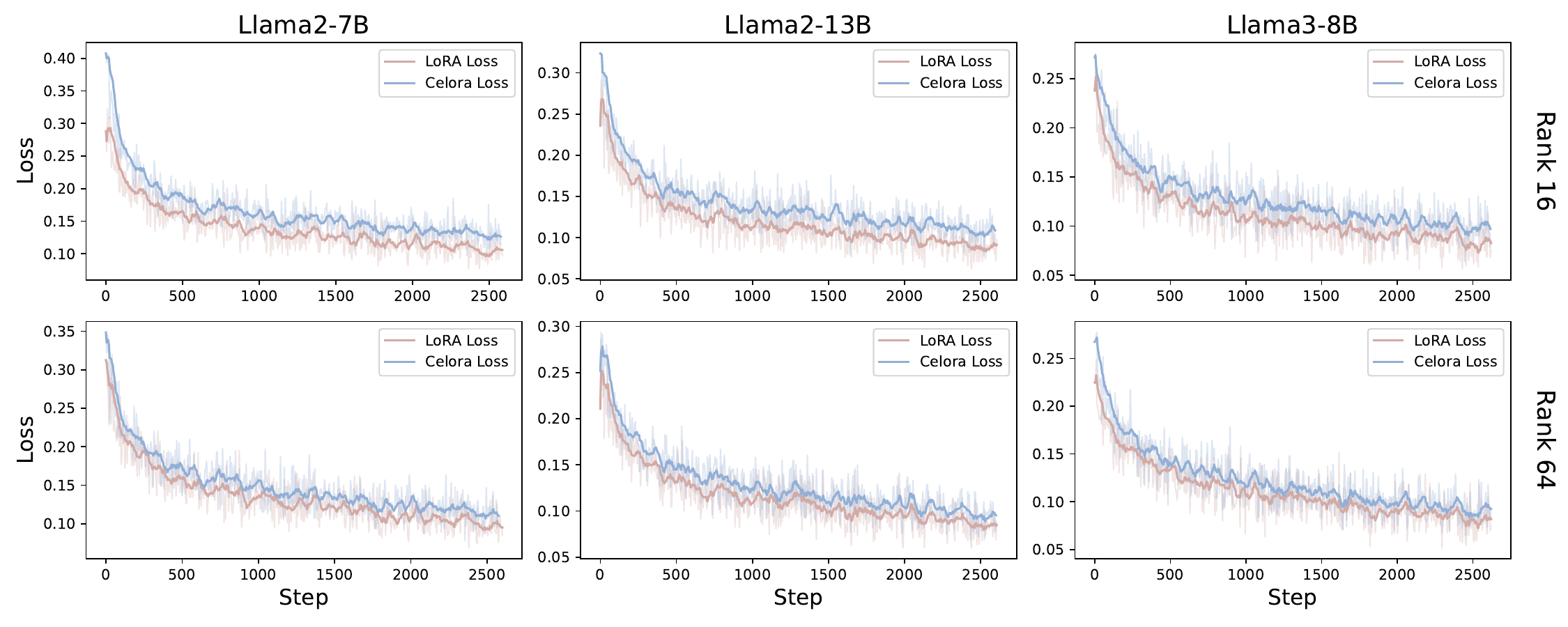}
  \vspace{-5mm}
  \caption{Loss curve of commonsense reasoning fine-tune task. Each row in the figure corresponds to a different trainable parameter setting, while each column represents base models: LLaMA2-7B/13B and LLaMA3.1-8B.}
  \label{fig:loss_curve}
\end{figure*}

In this set of experiments, we evaluate the convergence performance of \celora using two critical metrics: the accuracy achieved on each benchmark and the trajectory of the fine-tuning loss across training iterations. 
By monitoring these metrics, we aim to gain insights into how quickly and effectively \celora converges compared to LoRA.

\textbf{Accuracy}. 
We trained both \celora and LoRA under low-rank (LoRA rank of $16$, \celora rank of $14$) and high-rank (LoRA rank of $64$, \celora rank of $56$) configurations across two reasoning datasets for one epoch.
Table \ref{table:common_accuracy} and Table \ref{table:math_accuracy} summarize the results for the commonsense and arithmetic reasoning benchmarks.
The experimental outcomes demonstrate that, across all LoRA rank settings in both benchmarks, \celora achieves fine-tuning accuracy that is nearly identical to that of LoRA, with an average difference in results of 1.58\%.
These findings suggest that our approach has a negligible impact on the original LoRA fine-tuning accuracy.
The slight differences in accuracy between \celora and LoRA on the test sets can primarily be attributed to the scaling of \celora's rank, which was adjusted to ensure a fair experimental comparison.

\begin{table*}[!b]
    \caption{Performance comparison of LoRA and \celora on seven arithmetic reasoning tasks.}
    \label{table:math_accuracy}
    \vspace{0.5em}
    \centering
    \begin{adjustbox}{max width=\textwidth}
    \begin{tabular}{l l c c c c c c c c c}
        \toprule
        \textbf{Model} & \textbf{Method} & \textbf{Rank} & \textbf{MultiArith} & \textbf{GSM8K} & \textbf{AddSub} & \textbf{AQuA} & \textbf{SingleEq} & \textbf{SVAMP} & \textbf{MAWPS} & \textbf{Avg. $\uparrow$} \\
        \midrule

        \multirow{4}{*}{LLaMA3-8B} 
            & LoRA & 16 & 94.50 & 64.59 & 90.89 & 47.24 & 92.13 & 76.30 & 88.66 & 79.19 \\
            & \cellcolor{skyblue}\celora & \cellcolor{skyblue}14 & \cellcolor{skyblue}94.00 & \cellcolor{skyblue}62.09 & \cellcolor{skyblue}91.14 & \cellcolor{skyblue}44.88 & \cellcolor{skyblue}93.50 & \cellcolor{skyblue}75.00 & \cellcolor{skyblue}90.76 & \cellcolor{skyblue}78.77 \\
            & LoRA & 64 & 96.33 & 65.50 & 90.63 & 49.61 & 92.91 & 81.2 & 89.50 & 80.81 \\
            & \cellcolor{skyblue}\celora & \cellcolor{skyblue}56 & \cellcolor{skyblue}96.17 & \cellcolor{skyblue}62.02 & \cellcolor{skyblue}88.86 & \cellcolor{skyblue}47.64 & \cellcolor{skyblue}93.31 & \cellcolor{skyblue}77.10 & \cellcolor{skyblue}89.08 & \cellcolor{skyblue}79.17 \\
        \bottomrule
        
    \end{tabular}
    \end{adjustbox}
    
\end{table*}
\textbf{Loss curve}.
Figure \ref{fig:loss_curve} illustrates the loss curves of both \celora and LoRA under different rank settings across the three models on the commonsense reasoning fine-tuning task.
In each setting, \celora's loss curves nearly overlap with those of its LoRA counterparts, indicating similar convergence behaviors.
These results highlight the effectiveness of our method, empirically demonstrating that \celora can achieve nearly the same convergence capability as the original LoRA while potentially offering computational savings. 
The overlapping loss curves suggest that \celora does not introduce additional convergence challenges and maintains training stability comparable to LoRA.

\subsection{Computation Efficiency}

In these experiments, we measure \celora's training efficiency by comparing the average training step latency of a single-layer \celora with a single-layer LoRA. 
All experiments are conducted on a single \texttt{NVIDIA-HGX-H20-(96GB)} GPU to maintain consistent hardware conditions. 
For a fair comparison, both \celora and LoRA employ the same trainable rank of $64$. 
We run experiments on three different model weight sizes---$(8192, 8192)$, $(4096, 4096)$, and $(2048, 2048)$---using a fixed batch size of $16$ and a sequence length of $8192$. 
To measure average training step latency, each configuration is tested over 100 runs.
The first 10 iterations of each run are considered warmup and are excluded from latency measurements to mitigate initialization overhead.

\begin{figure*}[!t]
  \centering
    \includegraphics[width=\linewidth]{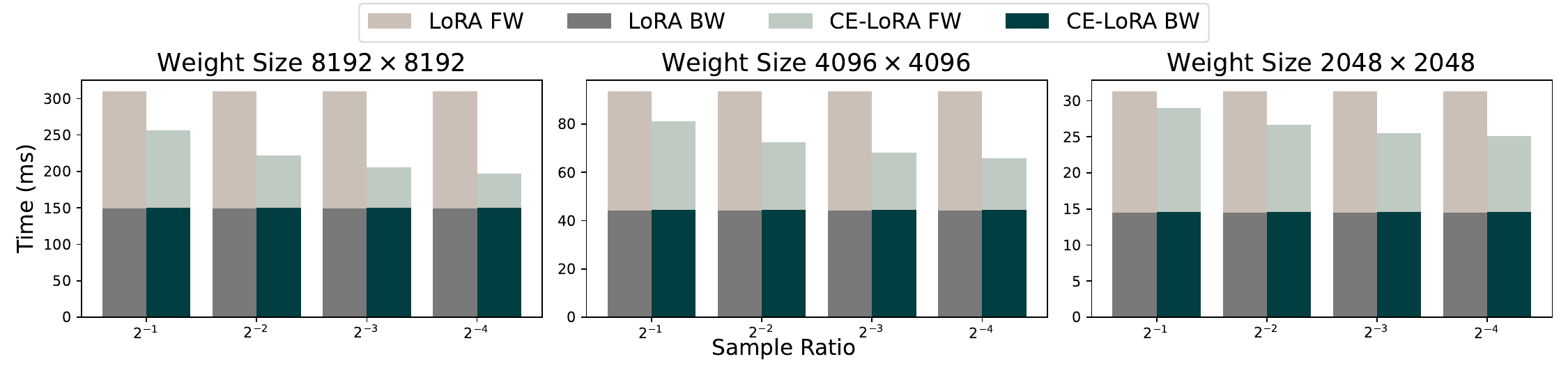}
  \vspace{-5mm}
  \caption{Comparison of training latency for \celora and LoRA at various sparsity levels (i.e., $\frac{1}{2}$, $\frac{1}{4}$, $\frac{1}{8}$, $\frac{1}{16}$) across three model shapes: $(8192, 8192)$, $(4096, 4096)$, and $(2048, 2048)$. \celora provides significant speedups in the backward pass, leading to a maximum of $36.3\%$ overall reduction in end-to-end training time compared to LoRA.}
  \label{fig:latency}
\end{figure*}

Figure~\ref{fig:latency} compares the results of LoRA and \celora with various sparsity levels and shows that \celora achieves a consistent reduction in overall training time, with a maximum of $36.3\%$ speedup. 
As illustrated, \celora’s forward pass latency closely matches that of LoRA's due to the unchanged forward logic of the frozen layer. 
However, in the backward pass, \celora outperforms LoRA by up to $3.39\times$ with some aggressive sampling rate.
The observed improvements in wall-clock speed are primarily attributed to two key factors:
(\underline{i}) \celora effectively reduces the theoretical floating-point operations required during backpropagation for frozen layers. 
(\underline{ii}) We developed specialized CUDA kernels tailored for low-rank computations inherent in \celora's backpropagation process, which optimize memory access patterns, resulting in enhanced computational efficiency and reduced latency.

\vspace{-0.25em}
\section{Related Works}
\vspace{-0.25em}
    \textbf{Large Language Models. }Since the transformer structure was proposed in the famous work \cite{vaswani2017attention} in 2017, it has shown great potential in various tasks, including reasoning, planning, machine translation, \textit{etc.}, and has become a popular choice in modern LLM designs, \textit{e.g.}, GPT \citep{radford2018improving,radford2019language,brown2020language}, OPT \citep{zhang2022opt}, LLaMA \citep{touvron2023llama,touvron2023Bllama,dubey2024llama}, BLOOM \citep{le2023bloom}, BERT \citep{devlin2018bert}, Falcon \citep{penedo2023refinedweb}, \textit{etc.} In general, the basic structure of a transformer block consists of a multi-head attention (MHA) module followed by a feed-forward network (FFN), combined with normalization and residual connections. Linear layers take up most of the trainable parameters in transformers and account for the expensive training and inference costs.

\textbf{Memory-Efficient Training Algorithms. }As the scale of LLM parameters grows, the memory consumption to train these models has become a bottleneck problem. Recent studies have proposed a series of works in order to reduce training-time memory consumption, enabling LLM researchers to effectively pre-train / fine-tune larger LLMs within constrained computational resources. \cite{houlsby2019parameter,pfeiffer2020adapterhub} fine-tune LLMs parameter-efficiently by adding trainable adapter layers, LoRA \citep{hu2021lora} reparameterizes linear layers in transformers with low-rank adapters, ReLoRA \citep{lialin2023relora} extends to pre-training tasks by accumulating LoRA updates, S$^2$FT \citep{yang2024sft} applies sparse structures, SLTrain \citep{han2024sltrain} combines low-rank and sparse structures. Besides the above parameter-efficient approaches, another line of works reduce the memory consumption for optimizer states by optimizing in periodically updated subspaces, including LISA \citep{pan2024lisa}, GaLore \citep{zhao2024galore}, GoLore \citep{he2024subspace} and \textsc{Flora} \citep{hao2024flora}. In addition, BackRazor \citep{jiang2022back} and PreBackRazor \citep{yu2024sheared} improve memory-efficiency by compressing the activation memory. Furthermore, quantization methods \citep{micikevicius2017mixed,dettmers2024qlora} that are orthogonal to the above approaches have shown nice compatibilities in memory cost reduction. 

\textbf{Computation-Efficient Training Algorithms. }Though not specially designed for computational efficiency, a lot of memory-efficient training algorithms, particularly  those belong to parameter-efficient fine-tuning (PEFT), can also reduce computational costs to some extent. On the other hand, the training throughput can also be improved by utilizing a larger batch size thanks to the reduced memory consumption \citep{zhu2024apollo}. However, the computational savings of these approaches are limited by precisely retaining the complete backward propagation process. Recently, \cite{woo2024dropbp} proposes DropBP, an approach orthogonal to PEFT that saves computation by strategically skip connections in backward propagation. Since some layers are dropped during backward propagation, corresponding parameters do not have gradients for update. To the best of our knowledge, this paper provides the \textit{first} approach to accelerate LoRA by employing structured sparsity to reduce the computational bottleneck in backward propagation without sacrificing memory-efficiency or model performance.


\vspace{-1em}
\section{Conclusion}

We propose \celora that saves computational FLOPs by approximated matrix multiplication and controls the compression error by a novel double LoRA technique and layer-wise compression ratios. While enjoying a 3.39 times of acceleration compared to LoRA, \celora theoretically converges at a rate of $\mathcal{O}(1/\sqrt{T})$ and shows comparable performance in our experiments.


\bibliographystyle{unsrt}
\bibliography{ref}

\clearpage
\appendix
\section{Missing Proofs}\label{app:proof}

In this section, we provide detailed proofs for Theorem \ref{thm:celora}. We first prove the following lemma.

\begin{lemma}\label{lm:m}
    Under Assumptions \ref{asp:proper}-\ref{asp:contractive}, if $\beta_1\in(0,1)$, it holds that
    \begin{align}
        \sum_{t=0}^T\mathbb{E}[\|\mathbf{m}^t-\nabla f(\mathbf{x}^t)\|_2^2]\le&\frac{2\|\mathbf{m}^0-\nabla f(\mathbf{x}^0)\|_2^2}{\beta_1}+\frac{4L^2}{\delta\beta_1^2}\sum_{t=1}^T\|\mathbf{x}^t-\mathbf{x}^{t-1}\|_2^2\nonumber\\
        &+\left(1-\frac{\delta}{2}\right)(1+6\beta_1)\sum_{t=1}^T\mathbb{E}[\|\nabla f(\mathbf{x}^t)\|_2^2]+6T\beta_1\sigma^2.\label{eq:lm-m}
    \end{align}
\end{lemma}
\begin{proof}
    According to the update of momentum, we have
    \begin{align}
        \mathbf{m}^{t}-\nabla f(\mathbf{x}^{t})=&(1-\beta_1)(\mathbf{m}^{t-1}-\nabla f(\mathbf{x}^{t}))+\beta_1(\hat{\mathbf{g}}^t-\nabla f(\mathbf{x}^t)).\nonumber
    \end{align}
    Taking expectation we have
    \begin{align}
        \mathbb{E}[\|\mathbf{m}^t-\nabla f(\mathbf{x}^t)\|_2^2]=&\mathbb{E}[\|(1-\beta_1)(\mathbf{m}^{t-1}-\nabla f(\mathbf{x}^t))+\beta_1(\mathbb{E}[\hat{\mathbf{g}}^t]-\nabla f(\mathbf{x}^t))\|_2^2]\nonumber\\
        &+\beta_1^2\mathbb{E}[\|\hat{\mathbf{g}}^t-\mathbb{E}[\hat{\mathbf{g}}^t]\|_2^2].\label{eq:pflm-m-1}
    \end{align}
    For the first term, applying Jensen's inequality yields
    \begin{align}
        &\mathbb{E}[\|(1-\beta_1)(\mathbf{m}^{t-1}-\nabla f(\mathbf{x}^t)+\beta_1(\mathbb{E}[\hat{\mathbf{g}}^t]-\nabla f(\mathbf{x}^t))\|_2^2]\nonumber\\
        \le&(1-\beta_1)\mathbb{E}[\|\mathbf{m}^{t-1}-\nabla f(\mathbf{x}^{t-1})-\nabla f(\mathbf{x}^t)+\nabla f(\mathbf{x}^{t-1})\|_2^2]+\beta_1\mathbb{E}[\|\mathbb{E}[\hat{\mathbf{g}}^t]-\nabla f(\mathbf{x}^t)\|_2^2].\label{eq:pflm-m-2}
    \end{align}
    By Young's inequality, we have
    \begin{align}
        \mathbb{E}[\|\mathbf{m}^{t-1}-\nabla f(\mathbf{x}^{t-1})-\nabla f(\mathbf{x}^t)+\nabla f(\mathbf{x}^{t-1})\|_2^2]\le&\left(1+\frac{\delta\beta_1}{2}\right)\mathbb{E}[\|\mathbf{m}^{t-1}-\nabla f(\mathbf{x}^{t-1})\|_2^2]\nonumber\\
        &+\left(1+\frac{2}{\delta\beta_1}\right)\mathbb{E}[\|\nabla f(\mathbf{x}^t)-\nabla f(\mathbf{x}^{t-1})\|_2^2].\label{eq:pflm-m-3}
    \end{align}
    For the second term, applying Cauchy's inequality yields
    \begin{align}
        \mathbb{E}[\|\hat{\mathbf{g}}^t-\mathbb{E}[\hat{\mathbf{g}}^t]\|_2^2]\le&3\mathbb{E}\|\hat{\mathbf{g}}^t-\mathbf{g}^t\|_2^2+3\mathbb{E}[\|\mathbf{g}^t-\nabla f(\mathbf{x}^t)\|_2^2]+3\mathbb{E}[\|\nabla f(\mathbf{x}^t)-\mathbb{E}[\hat{\mathbf{g}}^t]\|_2^2]\nonumber\\
        \le&6(1-\delta)\mathbb{E}[\|\nabla f(\mathbf{x}^t)\|_2^2]+3(2-\delta)\sigma^2,\label{eq:pflm-m-4}
    \end{align}
    where the last inequality uses Assumption \ref{asp:stochastic} and \ref{asp:contractive}.
    Applying \eqref{eq:pflm-m-2}\eqref{eq:pflm-m-3}\eqref{eq:pflm-m-4} to \eqref{eq:pflm-m-1} and using Assumption \ref{asp:smoothness} and \ref{asp:contractive}, we obtain
    \begin{align}
    \mathbb{E}[\|\mathbf{m}^t-\nabla f(\mathbf{x}^t)\|_2^2]\le&\left(1-\beta_1\left(1-\frac{\delta}{2}\right)\right)\mathbb{E}[\|\mathbf{m}^{t-1}-\nabla f(\mathbf{x}^{t-1})\|_2^2]+\frac{2L^2}{\delta\beta_1}\mathbb{E}[\|\mathbf{x}^t-\mathbf{x}^{t-1}\|_2^2]\nonumber\\
    &+(\beta_1+6\beta_1^2)(1-\delta)\mathbb{E}[\|\nabla f(\mathbf{x}^t)\|_2^2]+3(2-\delta)\beta_1^2\sigma^2.\label{eq:pflm-m-5}
    \end{align}
    Summing \eqref{eq:pflm-m-5} for $t=1,2,\cdots,T$ yields \eqref{eq:lm-m}.
\end{proof}

Now we are ready to prove Theorem \ref{thm:celora}. We first restate the theorem below in Theorem \ref{thm:celora-restate}.

\begin{theorem}\label{thm:celora-restate}
    Under Assumptions \ref{asp:proper}-\ref{asp:contractive}, if $\beta_1\in(0,\delta/(24-12\delta))$ and $\eta\le\min\{1/2L,\sqrt{(\delta\beta_1^2)/(8L^2)}\}$, CeLoRA with momentum SGD converges as
    \begin{align}
        \frac{1}{T+1}\sum_{t=0}^T\mathbb{E}[\|\nabla f(\mathbf{x}^t)\|_2^2]\le&\frac{4[f(\mathbf{x}^0)-\inf_{\mathbf{x}}f(\mathbf{x})]}{\delta\eta(T+1)}+\frac{4\|\mathbf{m}^0-\nabla f(\mathbf{x}^0)\|_2^2]}{\delta\beta_1(T+1)}+\frac{12\beta_1\sigma^2}{\delta}.\label{eq:thm-restate}
    \end{align}
\end{theorem}
\begin{proof}
    By Assumption \ref{asp:smoothness}, we have
    \begin{align}
        f(\mathbf{x}^{t+1})-f(\mathbf{x}^t)\le&\langle\nabla f(\mathbf{x}^t),\mathbf{x}^{t+1}-\mathbf{x}^t\rangle+\frac{L}{2}\|\mathbf{x}^{t+1}-\mathbf{x}^t\|_2^2\nonumber\\
        =&\left\langle\frac{\mathbf{m}^t}{2},\mathbf{x}^{t+1}-\mathbf{x}^t\right\rangle+\left\langle\nabla f(\mathbf{x}^t)-\frac{\mathbf{m}^t}{2},\mathbf{x}^{t+1}-\mathbf{x}^t\right\rangle+\frac{L}{2}\|\mathbf{x}^{t+1}-\mathbf{x}^t\|_2^2\nonumber\\
        =&-\left(\frac{1}{2\eta}-\frac{L}{2}\right)\|\mathbf{x}^{t+1}-\mathbf{x}^t\|_2^2+\frac{\eta}{2}\|\nabla f(\mathbf{x}^t)-\mathbf{m}^t\|_2^2-\frac{\eta}{2}\|\nabla f(\mathbf{x}^t)\|_2^2.\label{eq:pfthm-1}
    \end{align}
    Taking expectation and summing \eqref{eq:pfthm-1} for $t=0,1,\cdots,T$ yields
    \begin{align}
        \inf_{\mathbf{x}}f(\mathbf{x})-f(\mathbf{x}^0)\le&\frac{\eta}{2}\sum_{t=0}^{T}\mathbb{E}[\|\nabla f(\mathbf{x}^t)-\mathbf{m}^t\|_2^2]-\left(\frac{1}{2\eta}-\frac{L}{2}\right)\sum_{t=0}^{T}\mathbb{E}[\|\mathbf{x}^{t+1}-\mathbf{x}^t\|_2^2]\nonumber\\
        &-\frac{\eta}{2}\sum_{t=0}^T\mathbb{E}[\|\nabla f(\mathbf{x}^t)\|_2^2].\label{eq:pfthm-2}
    \end{align}
    Applying Lemma \ref{lm:m} to \eqref{eq:pfthm-2} and noting that $\beta_1\in(0,\delta/(24-12\delta))$ implies $(1-\delta/2)(1+6\beta_1)\le1-\delta/4$, we obtain
    \begin{align}
        \frac{1}{T+1}\sum_{t=0}^T\mathbb{E}[\|\nabla f(\mathbf{x}^t)\|_2^2]\le&\frac{4[f(\mathbf{x}^0)-\inf_{\mathbf{x}}f(\mathbf{x})]}{\delta\eta(T+1)}+\frac{4\|\mathbf{m}^0-\nabla f(\mathbf{x}^0)\|_2^2}{\delta\beta_1(T+1)}+\frac{12\beta_1\sigma^2}{\delta}\nonumber\\
        &-\frac{4}{\delta\eta}\left(\frac{1}{2\eta}-\frac{L}{2}-\frac{2\eta L^2}{\delta\beta_1^2}\right)\sum_{t=0}^T\|\mathbf{x}^{t+1}-\mathbf{x}^t\|_2^2.\label{eq:pfthm-3}
    \end{align}
   Since $\eta\le\min\{1/2L,\sqrt{(\delta\beta_1^2)/(8L^2)}\}$ implies $1/(4\eta)\ge L/2$ and $1/(4\eta)\ge(2\eta L^2)/(\delta\beta_1^2)$, \eqref{eq:thm-restate} is a direct result of \eqref{eq:pfthm-3}.
\end{proof}




\end{document}